\documentclass{article}

\usepackage{subfiles}

\usepackage{microtype}
\usepackage{graphicx}
\usepackage{subfigure}
\usepackage{booktabs} 

\usepackage{hyperref}

\usepackage[accepted]{icml2019}

\usepackage{amsmath}
\graphicspath{{figures/}}
\DeclareGraphicsExtensions{.jpg, .pdf, .png}
\DeclareMathOperator*{\argmax}{arg\,max}	

\icmltitlerunning{Sensitivity of Deep Convolutional Networks to Gabor Noise}

\begin{document}

\twocolumn[
\icmltitle{Sensitivity of Deep Convolutional Networks to Gabor Noise}

\begin{icmlauthorlist}
\icmlauthor{Kenneth T. Co}{icl}
\icmlauthor{Luis Mu\~noz-Gonz\'alez}{icl}
\icmlauthor{Emil C. Lupu}{icl}
\end{icmlauthorlist}

\icmlaffiliation{icl}{Department of Computing, Imperial College London, United Kingdom}
\icmlcorrespondingauthor{Kenneth T. Co}{k.co@imperial.ac.uk}
\icmlcorrespondingauthor{Luis Mu\~noz-Gonz\'alez}{l.munoz@imperial.ac.uk}
\icmlcorrespondingauthor{Emil C. Lupu}{e.c.lupu@imperial.ac.uk}

\vskip 0.3in
]

\printAffiliationsAndNotice{}

\section*{\normalsize Abstract}
Deep Convolutional Networks (DCNs) have been shown to be sensitive to Universal Adversarial Perturbations (UAPs): input-agnostic perturbations that fool a model on large portions of a dataset. These UAPs exhibit interesting visual patterns, but this phenomena is, as yet, poorly understood. Our work shows that visually similar procedural noise patterns also act as UAPs. In particular, we demonstrate that different DCN architectures are sensitive to Gabor noise patterns. This behaviour, its causes, and implications deserve further in-depth study.

\section{Introduction}
\begin{figure}[t]
	\centering
	\includegraphics[clip, width = 3.09in]{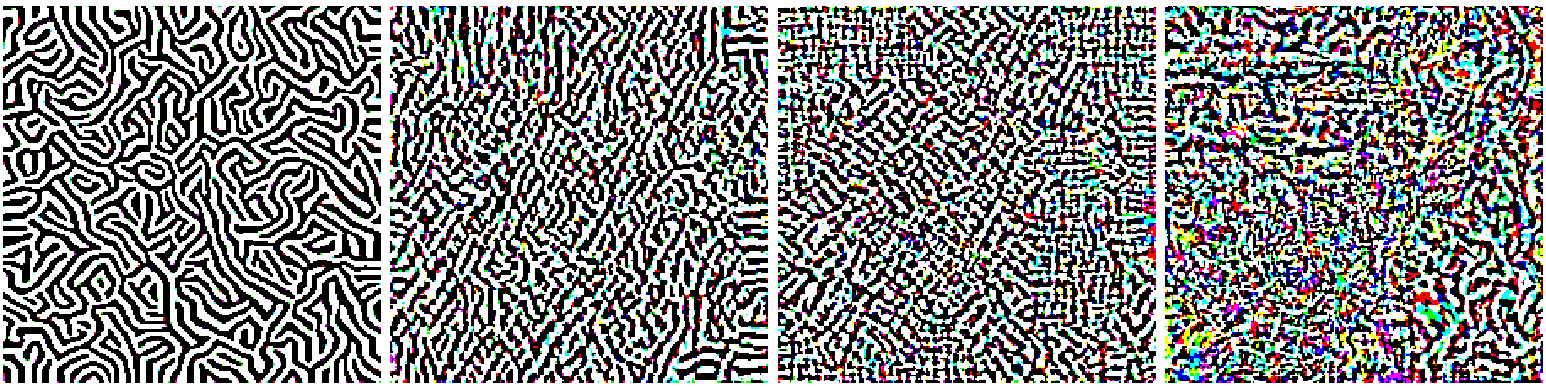}
	\caption{UAPs generated for VGG-19 targeting specific layers using singular vector method \cite{khrulkov2018art}.}
	\label{fig:1_uap_khrulkov}
\end{figure}

Deep Convolutional Networks (DCNs) have enabled deep learning to become one the primary tools for computer vision tasks. However, adversarial examples--slightly altered inputs that change the model's output--have raised concerns on their reliability and security. Adversarial perturbations can be defined as the noise patterns added to natural inputs to generate adversarial examples. Some of these perturbations are \emph{universal}, i.e. the same pattern can be used to fool the classifier on a large fraction of the tested dataset \cite{moosavi2017universal, khrulkov2018art}. As shown in Fig.~\ref{fig:1_uap_khrulkov}, it is interesting to observe that such Universal Adversarial Perturbations (UAPs) for DCNs contain structure in their noise patterns.

\begin{figure}[t]
	\centering
	\includegraphics[clip, width = 3.09in]{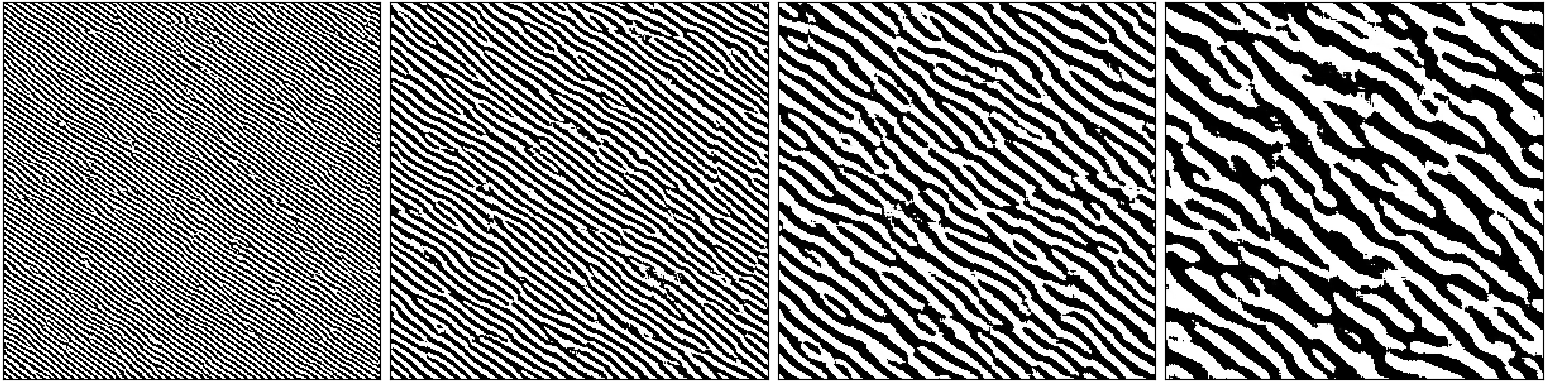}
	\caption{Gabor noise with normalized variance spectrums \cite{neyret2016understanding} and decreasing frequency from left to right.}
	\vspace{-0.1in}
	\label{fig:1_procedural}
\end{figure}

Results from \cite{co2018procedural} together with our results here suggest that DCNs are sensitive to procedural noise perturbations, and more specifically here to Gabor noise. Existing UAPs have some visual similarities with Gabor noise as in Figure~\ref{fig:1_procedural}. Convolutional layers induce a prior on DCNs to learn local spatial information \cite{goodfellow2016book}, and DCNs trained on natural image datasets, such as ImageNet, learn convolution filters that are similar in appearance to Gabor kernels and colour blobs \cite{yosinski2014transferable, olah2017feature}. Gabor noise is a convolution between a Gabor kernel\footnote[2]{A kernel (or filter) in image processing refers to a mask or small matrix used for image convolution.} and a sparse white noise. Thus, we hypothesize that DCNs are sensitive to Gabor noise, as it exploits specific features learned by the convolutional filters. 

In this paper we demonstrate the sensitivity of 3 different DCN architectures (\emph{Inception v3}, \emph{ResNet-50}, and \emph{VGG-19}), to Gabor noise on the ImageNet image classification task. We empirically observed that even random Gabor noise patterns can be effective to generate UAPs. Understanding this behaviour is important, as the generation and injection of Gabor noise is computationally inexpensive and, therefore, can become a threat to the security and reliability of DCNs.



\section{Background}
Compared to standard adversarial examples, UAPs reveal more general features that the DCN is sensitive to. In contrast, adversarial perturbations generated for specific inputs, though less detectable in many cases, can ``overfit'' and evade only on inputs they were generated for \cite{zhou2018transferable}. Previous approaches to generate UAPs use knowledge of the model's learned parameters. \citet{moosavi2017universal} use the DeepFool algorithm \cite{moosavi2016deepfool} iteratively over a set of images to construct a UAP. A different approach is proposed in \cite{mopuri2018nag}, where UAPs are computed using Generative Adversarial Nets (GANs). 

\citet{khrulkov2018art} proposed the singular vector method to generate UAPs targeting specific layers of DCNs, learning a perturbation $s$ that maximises the $L_p$-norm of the differences in the activations for that specific layer, $f_i$: 
$$\argmax_{s} \Vert f_i(x) - f_i(x + s) \Vert_{p} \\, \quad \Vert s \Vert_{q} = \varepsilon $$ 
where the $L_q$-norm of $s$ is constrained to $\varepsilon$. This can approximated using the Jacobian for that layer: 
$$\Vert f_i(x) - f_i(x + s) \Vert_{p} \approx \Vert J_i(x) \cdot s \Vert_{p}. $$
The solution $s$ that maximizes this is the $(p, q)$-singular vector can be computed with the power method \cite{boyd1974power}. Then, $s$ is effective to generate UAPs targeting a specific layer in the DCN. The solutions obtained with this method for the first layers of DCNs (see Fig.~\ref{fig:1_uap_khrulkov}) resemble the Gabor noise patterns shown in Fig.~\ref{fig:1_procedural}. 

However none of these works highlight the interesting visual patterns that manifest from these UAPs. In contrast, we show that procedural noise can generate UAPs targeting DCNs in a systematic and efficient way.

\section{Gabor Noise}
Gabor noise is the convolution of a sparse white noise and a Gabor kernel, making it a type of \emph{Sparse Convolution Noise} \cite{lagae2009procedural, lagae2010survey}. The Gabor kernel $g$ with parameters $\{\kappa, \sigma, \lambda, \omega\}$ is the product of a circular Gaussian and a harmonic function
$$ g(x, y) = \kappa e^{-\pi \sigma^2 (x^2 + y^2)} \cos{[2 \pi \lambda (x \cos \omega + y \sin \omega)]} $$
where $\kappa$ and $\sigma$ are the magnitude and width of the Gaussian, and $\lambda$ and $\omega$ are the frequency and orientation of the Harmonic \cite{lagae2010survey}. The value of the Gabor noise at point $(x, y)$ is given by
$$ G(x, y) = \sum_i w_i \, g(x - x_i, y - y_i; \kappa_i, \sigma_i, \lambda_i, \omega_i) $$
where $(x_i, y_i)$ are the coordinates of sparse random points and $w_i$ are random weights.

Gabor noise is an expressive noise function and has exponentially many parameterizations to explore. To simplify the analysis, we choose anisotropic Gabor noise, where the Gabor kernel parameters and weights are the same for each $i$. This results in noise patterns that have uniform orientation and thickness. We also normalize the variance spectrum of the Gabor noise using the algorithm in \cite{neyret2016understanding} to achieve min-max oscillations within the pattern.

\section{Experiments}

\begin{figure*}[t]
	\centering
	\includegraphics[width = 4.9in]{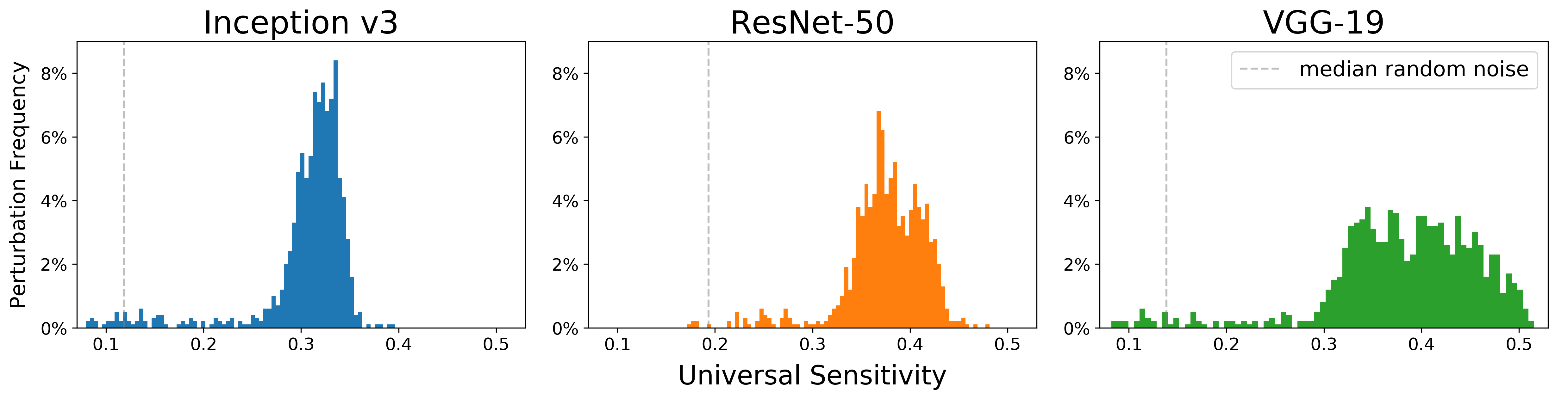}
	\includegraphics[width = 4.9in]{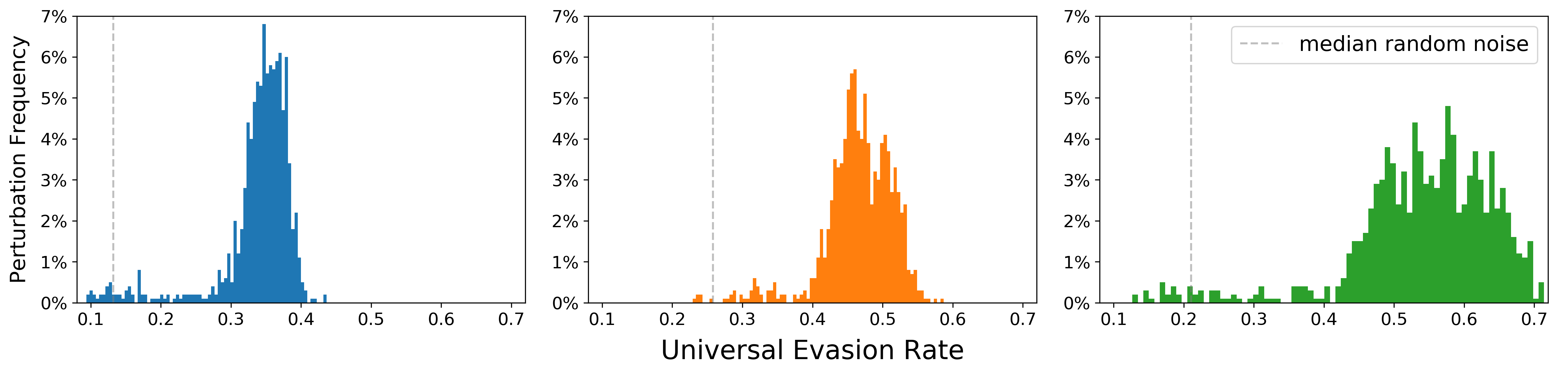}
	\caption{Histogram of 1,000 Gabor noise perturbations' (top) universal sensitivity and (bottom) universal evasion over 5,000 inputs.}
    	\label{fig:param_gba_histo}
\end{figure*}

For our experiments we use the validation set from the ILSVRC2012 ImageNet image classification task \cite{russakovsky2015imagenet} with 1,000 distinct categories. We use 3 pre-trained ImageNet DCN architectures from \texttt{keras.applications}: Inception v3 \cite{szegedy2016rethinking}, ResNet-50 \cite{he2016deep}, and VGG-19 \cite{simonyan2014very}.

Inception v3 take input images with dimensions $299 \times 299 \times 3$ while the other two networks take images with dimensions $224 \times 224 \times 3$. The kernel size $\kappa = 23$ is fixed so that the Gabor kernels will fill the entire image regardless of the distribution of points. The number of points $i$ distributed will be proportional to the image dimensions, which is independent of the Gabor kernel parameters. The resulting Gabor noise parameters we control are $\Theta = \{\sigma, \omega, \lambda\}$. We test the sensitivity of the models with 1,000 random Gabor noise perturbations generated from uniformly drawn parameters $\Theta$ with $\sigma, \lambda \in [1.5, 9]$ and $\omega \in [0, \pi]$.

We evaluate our Gabor noise on 5,000 random images from the validation set with an $\ell_{\infty}$ norm constraint of $\varepsilon = 12$ on the noise. The choice of $\frac{12}{256} \approx 0.047$ is consistent with other attacks on ImageNet-scale models with less than 5\% perturbation magnitude. To provide a baseline, we also measure the sensitivity of the models to 1,000 uniform random noise perturbations from $\{-\varepsilon, \varepsilon\}^{D \times D \times 3}$ where $D$ is the image's side length. This is useful for showing that the sensitivity to Gabor noise is not trivial.

\subsection{Metrics}
Given model output $f$, input $x \in X$, perturbation $s$, and small $\varepsilon > 0$, we define the \textbf{universal sensitivity} of a model on perturbation $s$ over $X$ as
$$ \frac{1}{\vert X \vert} \sum_{x \in X} \Vert f(x) - f(x + s) \Vert_{\infty}, \quad \Vert s \Vert_{\infty} = \varepsilon.$$
The norm constraint on $s$ ensures that the perturbation is small. For this paper, we choose $\infty$-norm as it is straightforward to impose for Gabor noise perturbations and is often used in the adversarial machine learning literature. For classification tasks, it is also useful to consider the \textbf{universal evasion} rate of a perturbation $s$ over $X$
$$ \frac{\vert \{x \in X : \argmax f(x) \neq \argmax f(x + s) \} \vert}{\vert X \vert}. $$
This corresponds to the definition that an adversarial perturbation is a small change that alters the predicted output label. Note that we are not interested in the ground truth labels for $x$ or $x + s$. We focus instead on how small changes to the input result in large changes to the model's \emph{original predictions}.

It is worth using both the universal sensitivity and the universal evasion metrics, as the former gives a continuous measure of the sensitivity, while the latter tells us on how much of the dataset that perturbation changes the decision of the model.

\subsection{Sensitivity to Gabor Noise}
Our results show that the order from least to most sensitive models are Inception v3, ResNet-50, and then VGG-19. This is not surprising as the validation accuracies of these models also appear in the same order. Overall, our experiments show that the three models are significantly more sensitive to the Gabor noise than random noise. The universal sensitivity and evasion rates of random noise have very small variance and their values are clustered around their medians. Table~\ref{table:param_quartiles_v19} shows how close the quartiles of random noise's are for VGG-19.

Inception v3 is also insensitive to random noise, but has a moderate sensitivity to Gabor noise. ResNet-50 appears to be more sensitive to the random noise than VGG-19, but VGG-19 is more sensitive to Gabor noise than ResNet-50. This implies that when comparing models higher sensitivity to one type of perturbation does not imply the same relationship for another type of perturbation.

The results in Fig.~\ref{fig:param_gba_histo} suggest that across the three models a random Gabor noise is likely to affect the model outputs on a third or more of the input dataset. From the histograms, the Gabor noise perturbations appear to centre around relatively high modes for both metrics. As an example, the first quartile of Gabor noise, as seen in Table~\ref{table:param_quartiles_v19}, has 49.3\% universal evasion, i.e. about 75\% of the Gabor noise perturbations change VGG-19's decision on about half or more of the input dataset. For the remainder of this analysis we focus on VGG-19 as it is the most sensitive model. Similar figures and statistics for the other two models are in the appendix.

\begin{figure*}[htb]
	\centering
	\includegraphics[width = 4.8in]{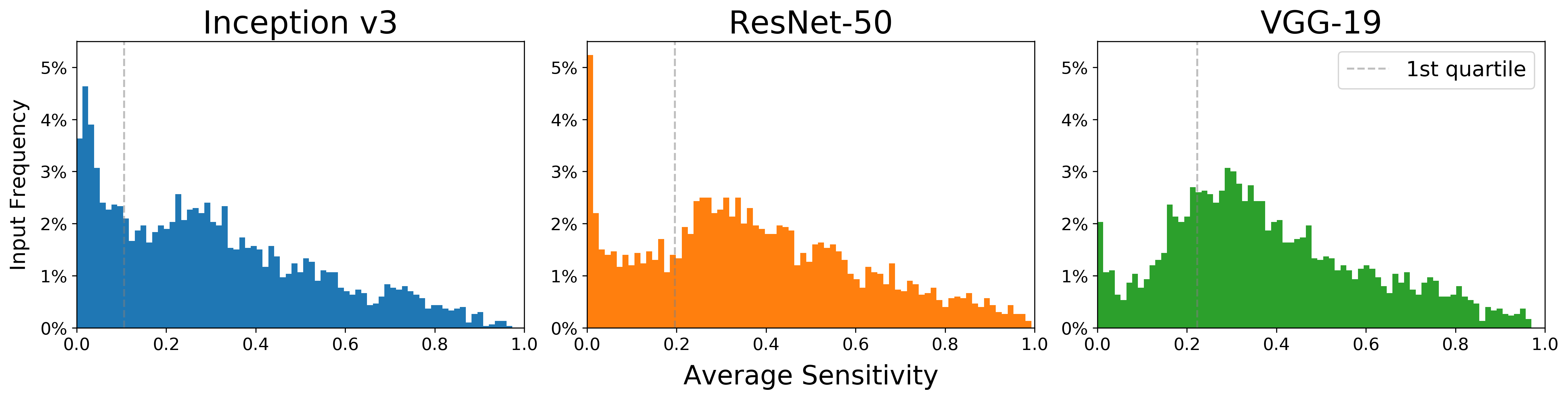}
	\includegraphics[width = 4.8in]{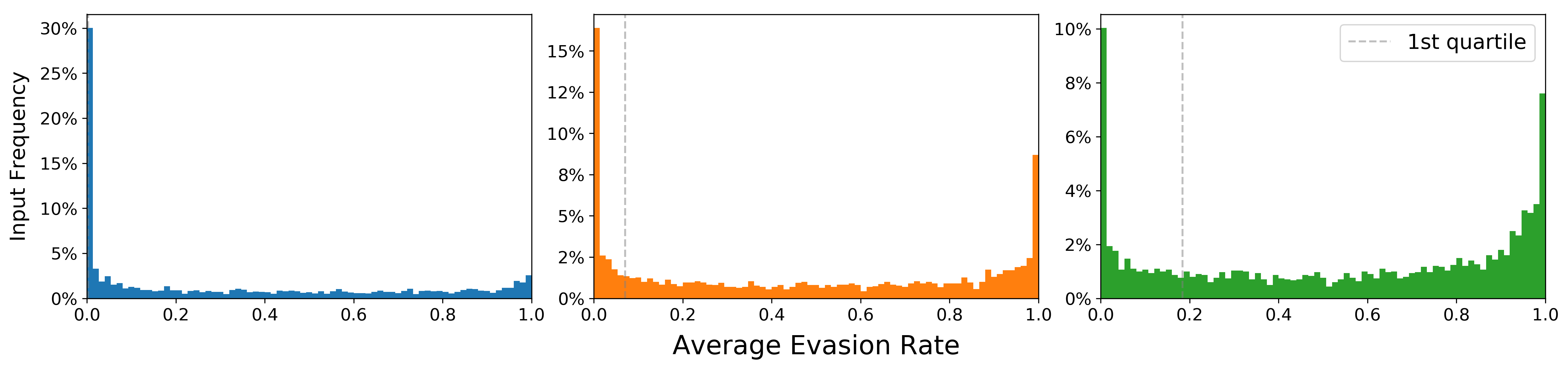}
    	\caption{Histogram of 5,000 inputs' (top) average sensitivity and (bottom) average evasion over 1,000 Gabor noise perturbations.}
        	\label{fig:img_histo}
\end{figure*}

\vspace{-0.2in}
\begin{table}[htp]
	\caption{Sensitivity (\%) metric quartiles of Gabor and random noise perturbations on VGG-19.}
	\label{table:param_quartiles_v19}
	\small
\begin{tabular*}{\linewidth}{@{\extracolsep{\fill}} l ccccc @{}}
	\\
	& \multicolumn{2}{c}{Universal Sensitivity} & \multicolumn{2}{c}{Universal Evasion}\\
	\cmidrule{2-3} \cmidrule{4-5}
	Quartile & Gabor & Random & Gabor & Random\\
	\midrule
	1st	& \textbf{34.2} & 13.8 & \textbf{49.3} & 20.8 \\
	2nd	& \textbf{39.1} & 13.9 & \textbf{55.5} & 21.0 \\
	3rd	& \textbf{43.7} & 13.9 & \textbf{61.5} & 21.3 \\
	\bottomrule
\end{tabular*}
\end{table}

\begin{figure}[htp]
	\centering
	\includegraphics[width = 3in]{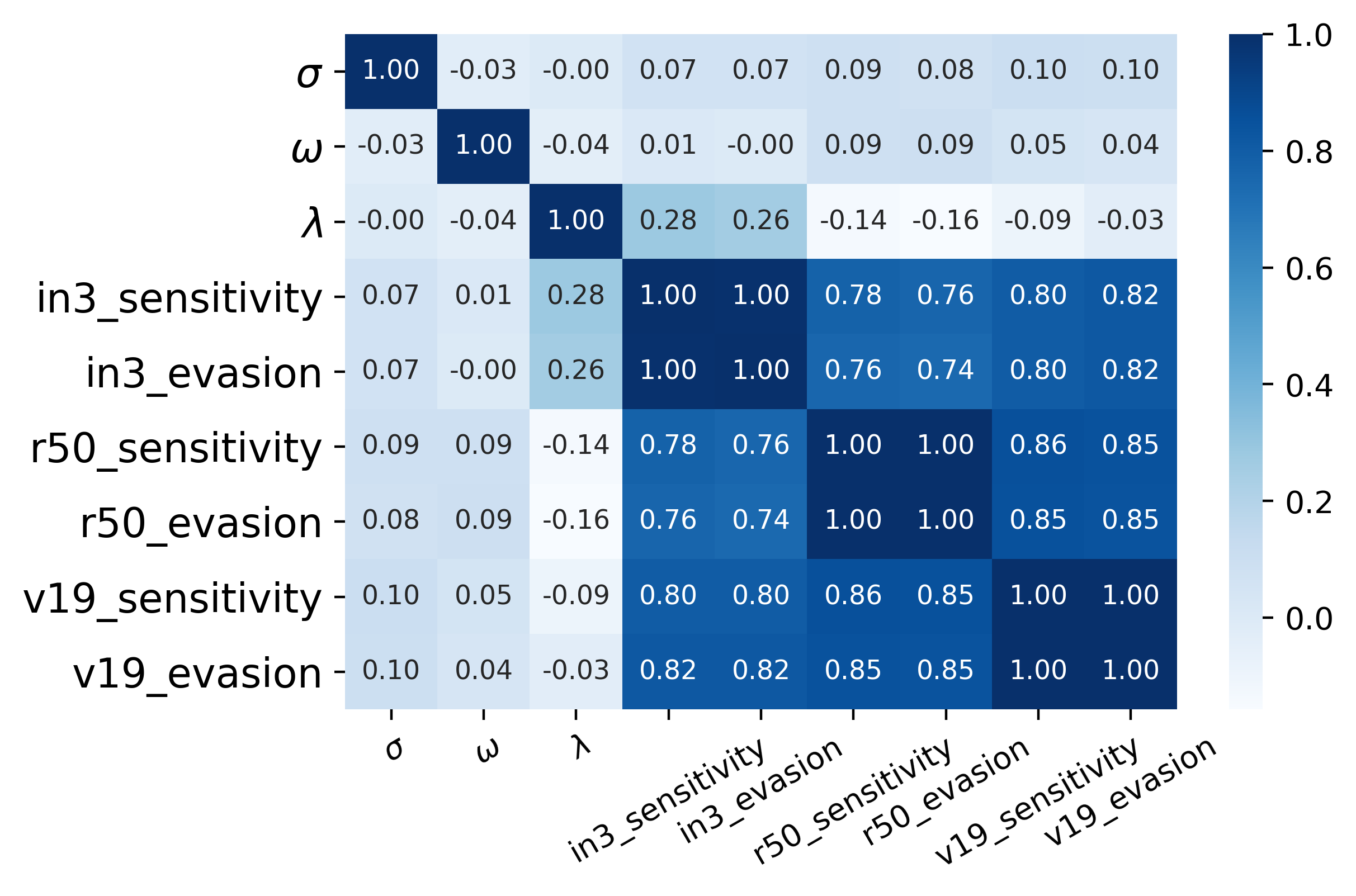}
	\caption{Correlation matrix of Gabor noise parameters and metrics for each model. Inception v3, ResNet-50, and VGG-19 are referred to as ``in3'', ``r50'', and ``v19'' respectively.} 
        	\label{fig:corr_mat}
\end{figure}

\textbf{``Best'' Parameters.} Taking the top 10 perturbations that VGG-19 is most sensitive to, we see that the other two models are also very sensitive to these noise patterns. The ranges of the universal evasion rate for these are 69.7\% to 71.4\% for VGG-19, 50.7\% to 53.4\% for ResNet-50, and 37.9\% to 39.4\% for Inception v3. These values are all above the 3rd quartile for each of these models, showing its generalizability to the other models.

In Fig.~\ref{fig:corr_mat} we see a strong correlation ($\geq$ 0.74) between the universal sensitivity and evasion rates across models. This further suggests that strong perturbations transfer across these models. We also see a weak correlation between $\lambda$ and the sensitivity and evasion rates for Inception v3, though there appears to be none between $\lambda$ and the sensitivity values for ResNet50.

The universal evasion rate of the perturbations appears to be insensitive to its Gaussian width $\sigma$ and orientation $\omega$. However, the sensitivity for small $\lambda < 0.3$ appears to fall below the average, suggesting that below a certain value the Gabor noise does not affect the model's decision. Interestingly, $\lambda$ corresponds to the width or thickness of the bands in the image. Examples of Gabor noise perturbations can be seen in the appendix.

\pagebreak
\textbf{Sensitivity of Inputs.} The model's sensitivity could vary across the input dataset, meaning that the model's predictions is stable on some inputs while more susceptible to small perturbations on others. To measure this, we look at the sensitivity of single inputs over all perturbations. 

Given a set of perturbations $s \in S$, we define the \textbf{average sensitivity} of a model on input $x$ over $S$ as
$$ \frac{1}{\vert S \vert} \sum_{s \in S} \Vert f(x) - f(x + s) \Vert_{\infty}, $$
and the \textbf{average evasion rate} on $x$ over $S$ as
$$ \frac{\vert \{s \in S : \argmax f(x) \neq \argmax f(x + s) \} \vert}{\vert S \vert}. $$

The bimodal distribution of the average evasion rate in Fig.~\ref{fig:img_histo} shows that for each model there are two large subsets of the data: One that is very sensitive and another that is very insensitive. The remaining data points are somewhat uniformly spread in the middle. Note that for Inception v3, there is a much larger fraction of data points whose prediction is not affected by Gabor perturbations. The distribution for the average sensitivity appears to have similar shape, but with more inputs in the 0-20\% range for Inception v3. The dataset is far less sensitive against random noise with upwards of 60\% of the dataset being insensitive to that noise across all models.

\section{Conclusion}
The results show that the tested DCN models are sensitive to Gabor noise for a large fraction of the inputs, even when the parameters of the Gabor noise are chosen at random. This hints that it may be representative of patterns learned at the earlier layers as Gabor noise appears visually similar to some UAPs targeting earlier layers in DCNs \cite{khrulkov2018art}.

This phenomenon has important implications on the security and reliability of DCNs, as it can allow attackers to craft inexpensive black-box attacks. On the defender's side, Gabor noise patterns can also be used to efficiently generate data for adversarial training to improve DCNs robustness. However, both the sensitivity exploited and the potential to mitigate it require a more in-depth understanding of the phenomena at play. In future work, it may be worth analyzing the sensitivity of hidden layer activations across different families of procedural noise patterns and to investigate techniques to reduce the sensitivity of DCNs to perturbations.

\section*{Acknowledgements}
Kenneth Co is partially supported by the Data Spartan research grant DSRD201801. Example code is available at \texttt{https://github.com/kenny-co/procedural-advml}

\raggedbottom


\fancyhead[LE, RO]{\slshape}
\bibliographystyle{icml2019}
\bibliography{icml2019}

\raggedbottom
\pagebreak

\appendix
\section{Sensitivity to Gabor Noise}
\begin{figure*}[htb]
	\centering
	\includegraphics[width = 5in]{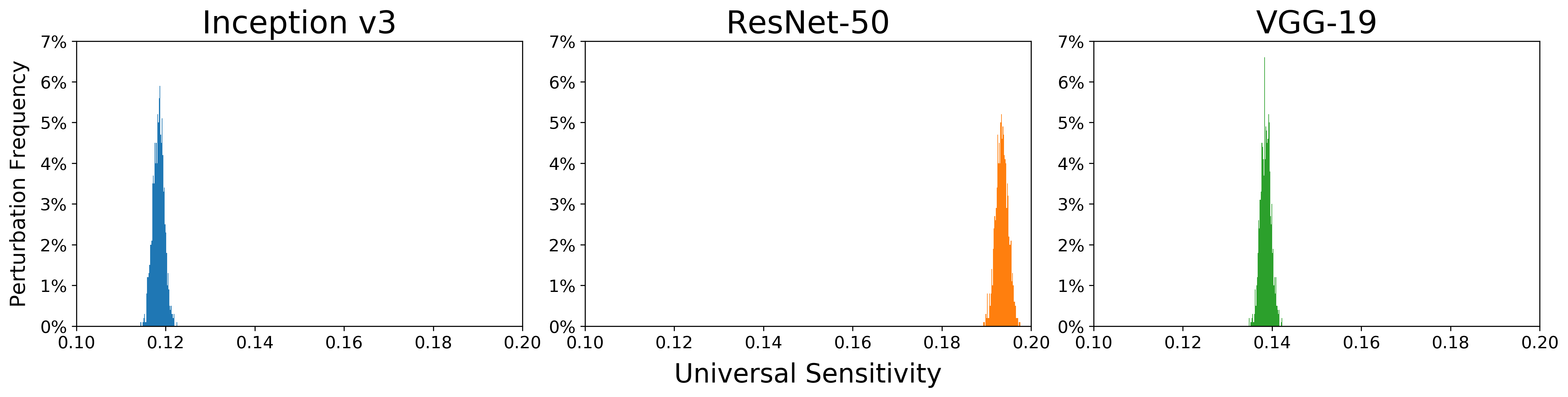}
	\includegraphics[width = 5in]{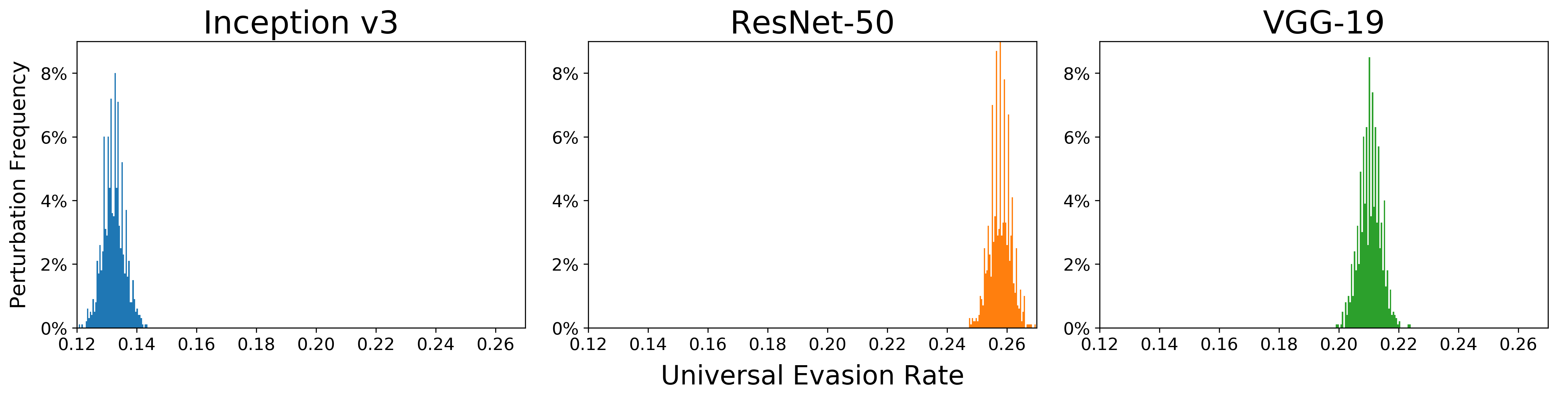}
	\caption{Histogram of 1,000 random noise perturbations based on (top) universal sensitivity and (bottom) universal evasion rate.}
    	\label{fig:param_ran_histo}
\end{figure*}

As seen in Figure~\ref{fig:param_ran_histo}, sensitivity metric values for random noise fall in a narrow range and are significantly smaller than the metric values of the Gabor noise. This is further shown when comparing the quartiles of the universal evasion and sensitivity in Tables \ref{table:param_ran_quartiles_in3} and \ref{table:param_ran_quartiles_r50}.

\begin{figure}[htb]
	\centering
	\includegraphics[trim = {0.15cm 0cm 0.05cm 0.1cm}, clip, width = 3.09in]{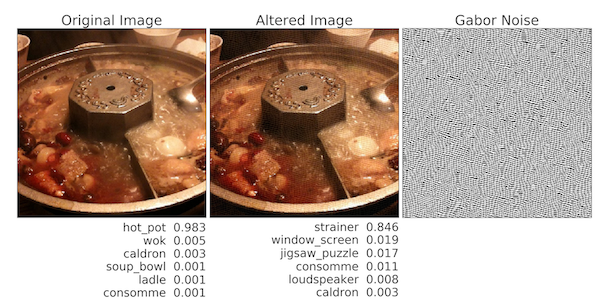}
	\includegraphics[trim = {0.15cm 0cm 0.05cm 0.1cm}, clip, width = 3.09in]{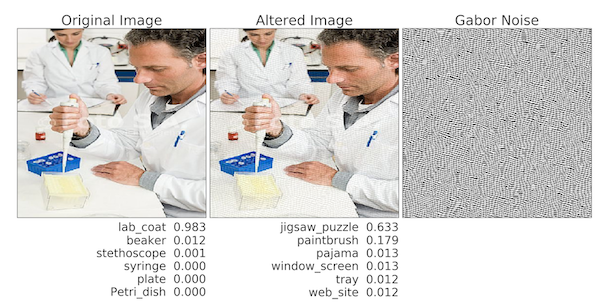}
	\includegraphics[trim = {0.15cm 0cm 0.05cm 0.1cm}, clip, width = 3.09in]{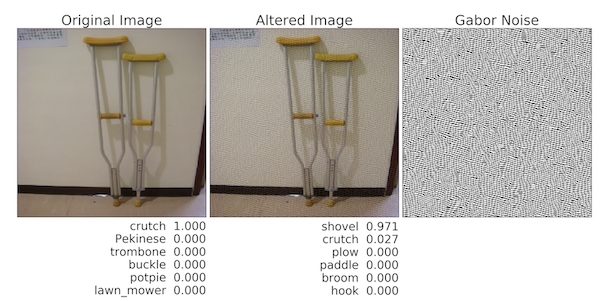}
	\caption{Gabor noise parameters $\Theta_{6} = \{4.78, 1.81, 2.93\}$ on Inception v3.}
	\label{fig:1}
\end{figure}

\begin{figure}[htb]
	\centering
	\includegraphics[trim = {0.15cm 0cm 0.05cm 0.1cm}, clip, width = 3.09in]{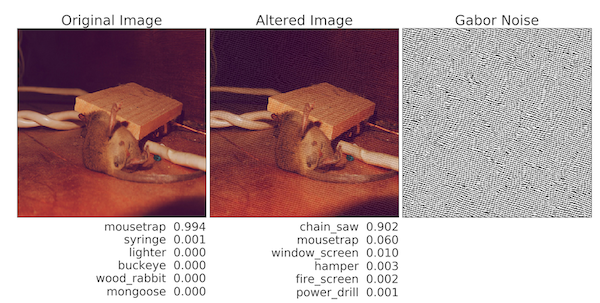}
	\includegraphics[trim = {0.15cm 0cm 0.05cm 0.1cm}, clip, width = 3.09in]{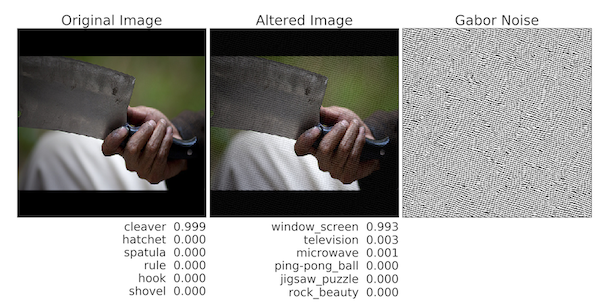}
	\includegraphics[trim = {0.15cm 0cm 0.05cm 0.1cm}, clip, width = 3.09in]{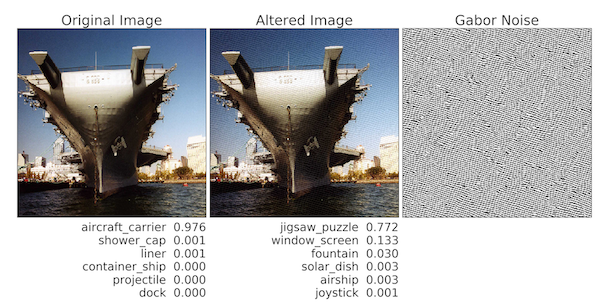}
	\caption{Gabor noise parameters $\Theta_{709} = \{7.92, 1.85, 3.12\}$ on Inception v3.}
	\label{fig:2}
\end{figure}

\begin{figure}[htb]
	\centering
	\includegraphics[trim = {0.15cm 0cm 0.05cm 0.1cm}, clip, width = 3.09in]{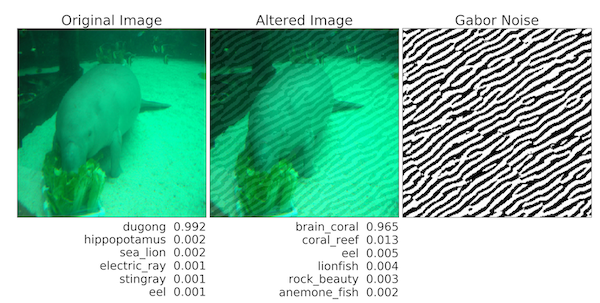}
	\includegraphics[trim = {0.15cm 0cm 0.05cm 0.1cm}, clip, width = 3.09in]{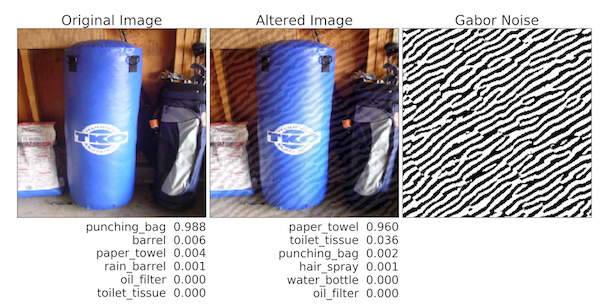}
	\includegraphics[trim = {0.15cm 0cm 0.05cm 0.1cm}, clip, width = 3.09in]{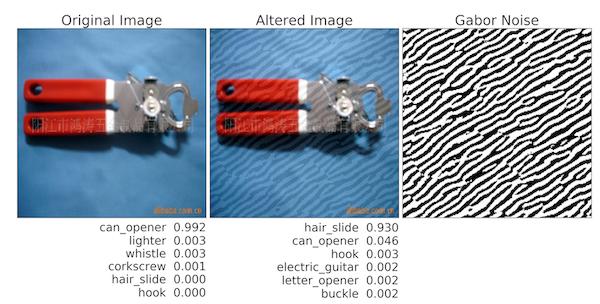}
	\includegraphics[trim = {0.15cm 0cm 0.05cm 0.1cm}, clip, width = 3.09in]{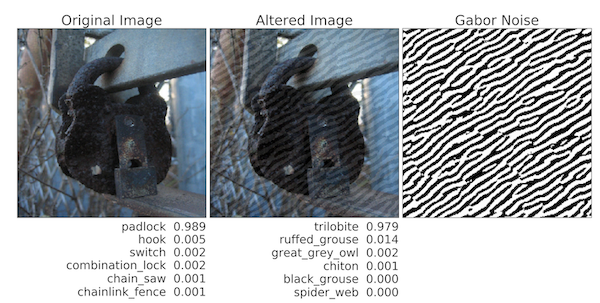}
	\caption{Gabor noise parameters $\Theta_{119} = \{6.69, 1.02, 8.45\}$ on ResNet-50.}
	\label{fig:3}
\end{figure}

\begin{figure}[htb]
	\centering
	\includegraphics[trim = {0.15cm 0cm 0.05cm 0.1cm}, clip, width = 3.09in]{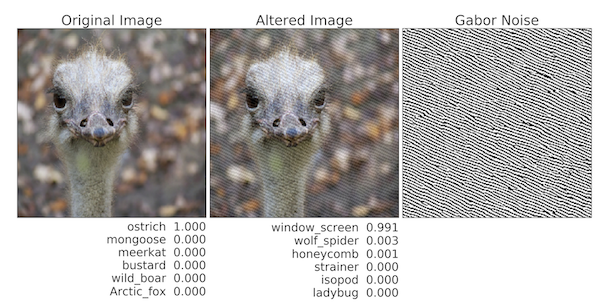}
	\includegraphics[trim = {0.15cm 0cm 0.05cm 0.1cm}, clip, width = 3.09in]{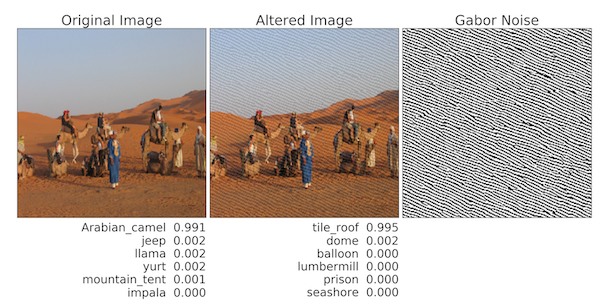}
	\includegraphics[trim = {0.15cm 0cm 0.05cm 0.1cm}, clip, width = 3.09in]{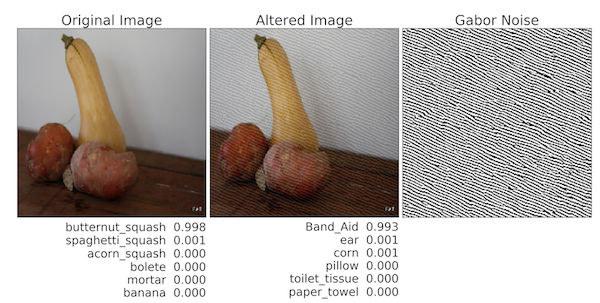}
	\caption{Gabor noise parameters $\Theta_{185} = \{6.10, 1.99, 3.46\}$ on ResNet-50.}
	\label{fig:4}
\end{figure}

\begin{figure}[htb]
	\centering
	\includegraphics[trim = {0.15cm 0cm 0.05cm 0.1cm}, clip, width = 3.09in]{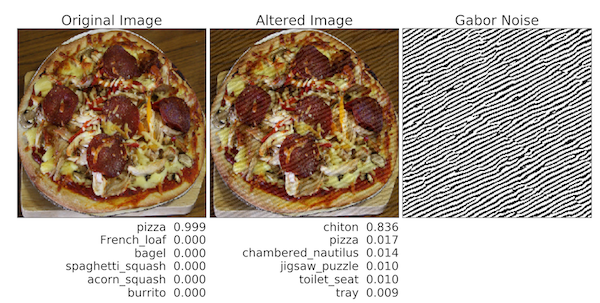}
	\includegraphics[trim = {0.15cm 0cm 0.05cm 0.1cm}, clip, width = 3.09in]{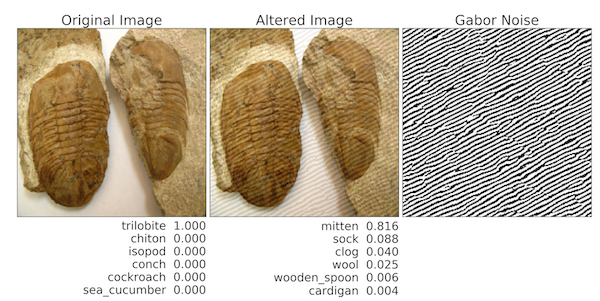}
	\includegraphics[trim = {0.15cm 0cm 0.05cm 0.1cm}, clip, width = 3.09in]{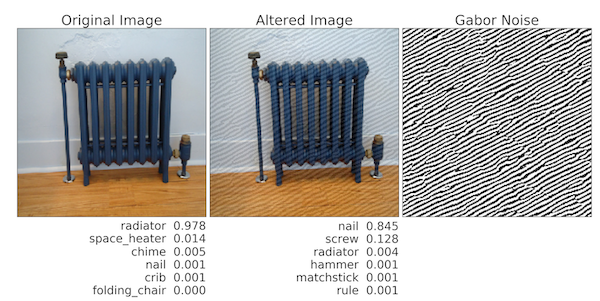}
	\includegraphics[trim = {0.15cm 0cm 0.05cm 0.1cm}, clip, width = 3.09in]{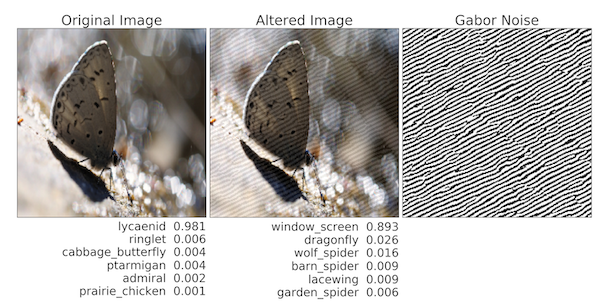}
	\includegraphics[trim = {0.15cm 0cm 0.05cm 0.1cm}, clip, width = 3.09in]{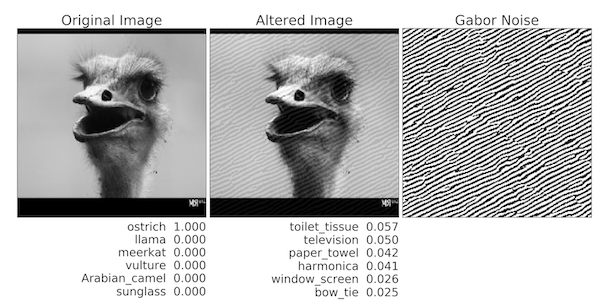}
	\caption{Gabor noise parameters $\Theta_{25} = \{6.29, 1.10, 4.86\}$ on VGG-19.}
	\label{fig:5}
\end{figure}

Figures~\ref{fig:1}, \ref{fig:2}, \ref{fig:3}, \ref{fig:4}, and \ref{fig:5} show some adversarial examples with the top perturbations.

\begin{table}[htp]
	\caption{Sensitivity (\%) metric quartiles of Gabor and random noise perturbations on Inception v3.}
	\label{table:param_ran_quartiles_in3}
\begin{tabular*}{\linewidth}{@{\extracolsep{\fill}} l ccccc @{}}
	\\
	& \multicolumn{2}{c}{Universal Sensitivity} & \multicolumn{2}{c}{Universal Evasion}\\
	\cmidrule{2-3} \cmidrule{4-5}
	Quartile & Gabor & Random & Gabor & Random\\
	\midrule
	1st	& \textbf{29.9} & 11.8 & \textbf{32.8} & 13.0 \\
	2nd	& \textbf{31.8} & 11.8 & \textbf{34.9} & 13.2 \\
	3rd	& \textbf{33.2} & 11.9 & \textbf{36.9} & 13.5 \\
	\bottomrule
\end{tabular*}
\end{table}
\begin{table}[htp]
	\caption{Sensitivity (\%) metric quartiles of Gabor and random noise perturbations on ResNet-50.}
	\label{table:param_ran_quartiles_r50}
\begin{tabular*}{\linewidth}{@{\extracolsep{\fill}} l ccccc @{}}
	\\
	& \multicolumn{2}{c}{Universal Sensitivity} & \multicolumn{2}{c}{Universal Evasion}\\
	\cmidrule{2-3} \cmidrule{4-5}
	Quartile & Gabor & Random & Gabor & Random\\
	\midrule
	1st	& \textbf{35.7} & 19.3 & \textbf{44.3} & 25.6 \\
	2nd	& \textbf{37.7} & 19.3 & \textbf{46.8} & 25.8 \\
	3rd	& \textbf{40.4} & 19.4 & \textbf{50.1} & 26.0 \\
	\bottomrule
\end{tabular*}
\end{table}

\vfill\eject
\raggedbottom


\section{Sensitivity of Inputs}
\begin{figure*}[htb]
	\centering
	\includegraphics[width = 5in]{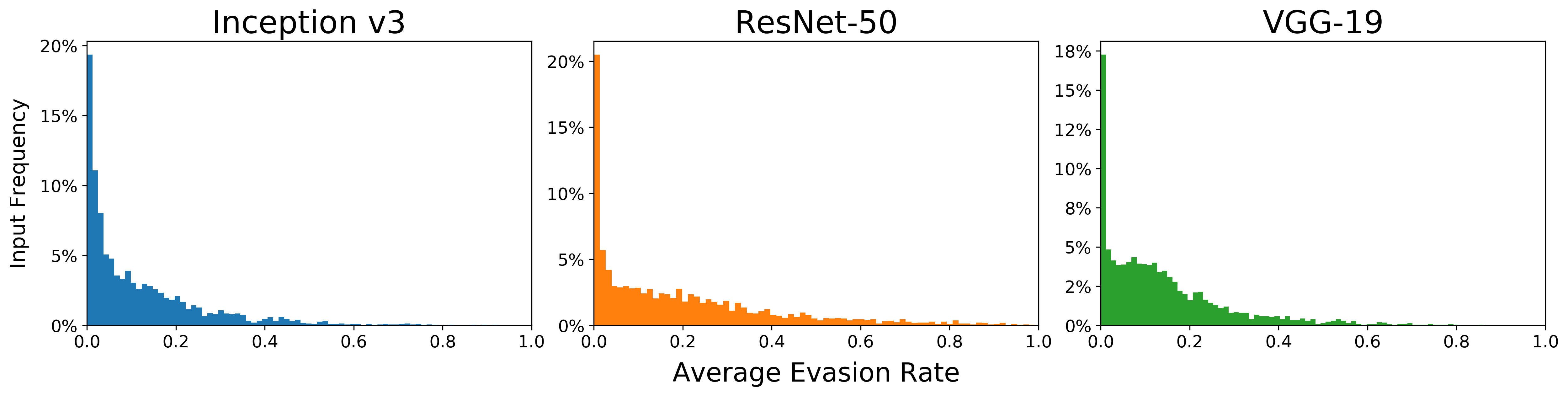}
	\includegraphics[width = 5in]{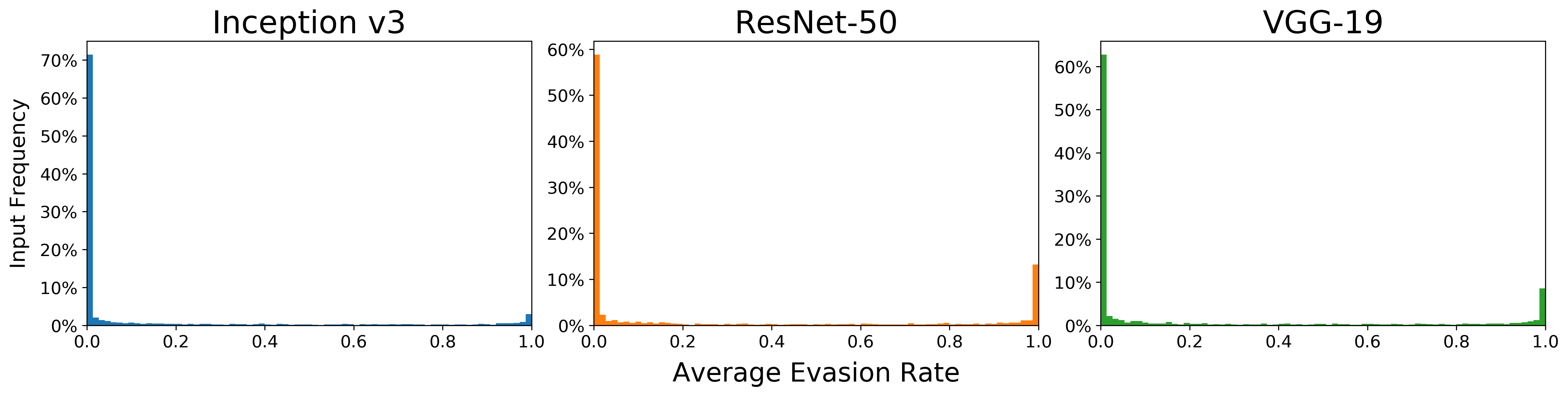}
	\caption{Histogram of 5,000 inputs based on (top) average sensitivity and (bottom) average evasion rate over random noise perturbations.}
    	\label{fig:img_ran_histo}
\end{figure*}

Large part of the input dataset is insensitive to random noise as shown in Tables~\ref{table:1}, \ref{table:2}, \ref{table:3} and Figure~\ref{fig:img_ran_histo}. With about 60\% of the dataset on having near 0\% average evasion over the random noise perturbations for all three models.

\begin{table}[htp]
	\caption{Sensitivity (\%) metric quartiles of input data over perturbations on Inception v3.}
	\label{table:1}
\begin{tabular*}{\linewidth}{@{\extracolsep{\fill}} l ccccc @{}}
	\\
	& \multicolumn{2}{c}{Average Sensitivity} & \multicolumn{2}{c}{Average Evasion}\\
	\cmidrule{2-3} \cmidrule{4-5}
	Quartile & Gabor & Random & Gabor & Random\\
	\midrule
	1st	& \textbf{10.6} & 1.8 & \textbf{0.3} & 0.0 \\
	2nd	& \textbf{26.8} & 6.7 & \textbf{19.8} & 0.0 \\
	3rd	& \textbf{45.1} & 17.0 & \textbf{65.6} & 4.1 \\
	\bottomrule
\end{tabular*}
\end{table}
\begin{table}[htp]
	\caption{Sensitivity (\%) metric quartiles of input data over perturbations on ResNet-50.}
	\label{table:2}
\begin{tabular*}{\linewidth}{@{\extracolsep{\fill}} l ccccc @{}}
	\\
	& \multicolumn{2}{c}{Average Sensitivity} & \multicolumn{2}{c}{Average Evasion}\\
	\cmidrule{2-3} \cmidrule{4-5}
	Quartile & Gabor & Random & Gabor & Random\\
	\midrule
	1st	& \textbf{19.6} & 2.3 & \textbf{7.0} & 0.0 \\
	2nd	& \textbf{34.8} & 13.3 & \textbf{45.1} & 0.0 \\
	3rd	& \textbf{53.9} & 29.2 & \textbf{84.6} & 53.7 \\
	\bottomrule
\end{tabular*}
\end{table}
\begin{table}[htp]
	\caption{Sensitivity (\%) metric quartiles of input data over perturbations on VGG-19.}
	\label{table:3}
\begin{tabular*}{\linewidth}{@{\extracolsep{\fill}} l ccccc @{}}
	\\
	& \multicolumn{2}{c}{Average Sensitivity} & \multicolumn{2}{c}{Average Evasion}\\
	\cmidrule{2-3} \cmidrule{4-5}
	Quartile & Gabor & Random & Gabor & Random\\
	\midrule
	1st	& \textbf{22.3} & 3.1 & \textbf{18.4} & 0.0 \\
	2nd	& \textbf{34.2} & 10.3 & \textbf{60.2} & 0.0 \\
	3rd	& \textbf{52.2} & 19.6 & \textbf{90.0} & 24.5 \\
	\bottomrule
\end{tabular*}
\end{table}

\begin{figure*}[htb]
	\centering
	\includegraphics[width = 5in]{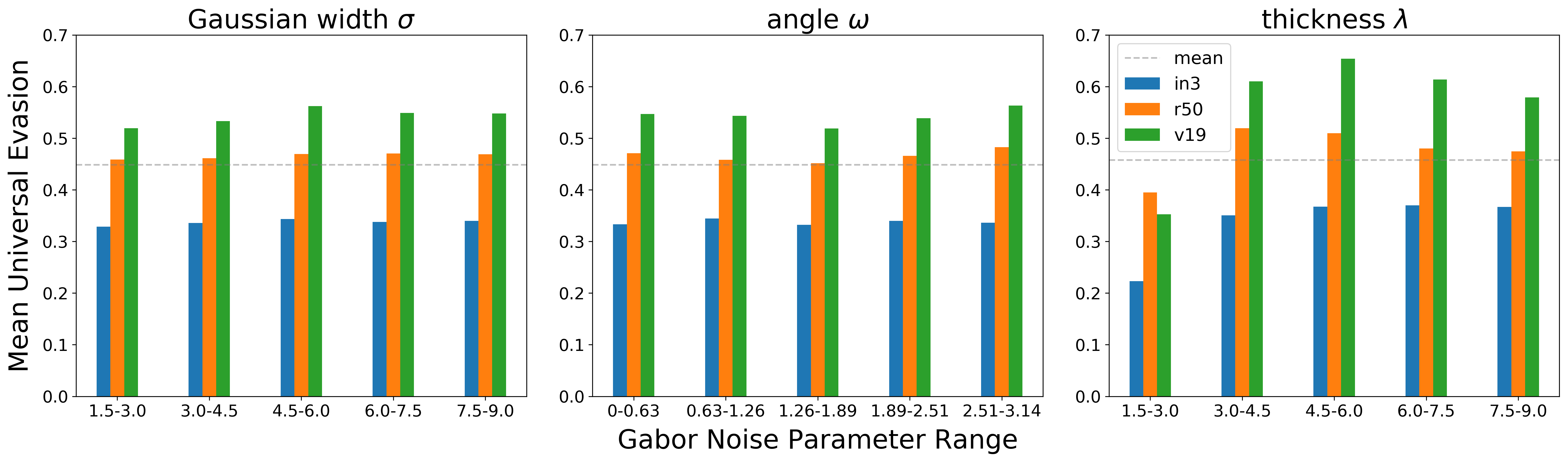}
	\caption{Mean universal evasion rate of Gabor noise perturbations grouped according to parameter values.}
	\label{fig:param_gba_comparison}
\end{figure*}

\end{document}